\pdfoutput=1

\documentclass[11pt]{article}

\usepackage[final]{EMNLP2022}

\usepackage{times}
\usepackage{latexsym}
\usepackage{graphicx}
\usepackage{subcaption}
\usepackage{caption}
\usepackage{comment}
\usepackage[T1]{fontenc}

\usepackage[utf8]{inputenc}

\usepackage{microtype}

\title{English Contrastive Learning Can Learn Universal Cross-lingual\\ Sentence Embeddings
}

\author{
  Yau-Shian Wang \quad Ashley Wu \quad Graham Neubig\\
  Carnegie Mellon University \\
  {\tt king6101@gmail.com  wangchew@andrew.cmu.edu}\\
  {\tt gneubig@cs.cmu.edu }\\
  }

\begin{document}
\maketitle
\begin{abstract}
Universal cross-lingual sentence embeddings map semantically similar cross-lingual sentences into a shared embedding space. Aligning cross-lingual sentence embeddings usually requires supervised cross-lingual parallel sentences. In this work, we propose mSimCSE, which extends SimCSE~\cite{gao2021simcse} to multilingual settings and reveal that contrastive learning on English data can surprisingly learn high-quality universal cross-lingual sentence embeddings without any parallel data.
In unsupervised and weakly supervised settings, mSimCSE significantly improves previous sentence embedding methods on cross-lingual retrieval and multilingual STS tasks. 
The performance of unsupervised mSimCSE is comparable to fully-supervised methods in retrieving low-resource languages and multilingual STS.
The performance can be further enhanced when cross-lingual NLI data is available.~\footnote{Our code is publicly available at \url{https://github.com/yaushian/mSimCSE}.}

\end{abstract}

\section{Introduction}
Universal cross-lingual sentence embeddings map the sentences from multiple languages into a shared embedding space, where semantically similar sentences across languages are close to each other.
These embeddings have a wide spectrum of applications such as multi-lingual document retrieval~\cite{Artetxe_2019,https://doi.org/10.48550/arxiv.2010.06467}, multi-lingual question answering~\cite{asai-etal-2021-xor,NEURIPS2021_3df07fda,kumar2022mucot}, unsupervised machine translation~\cite{NEURIPS2020_1763ea5a}, and zero-shot transfer learning~\cite{phang-etal-2020-english}.

\begin{figure}
\centering
\subcaptionbox{XLM-R without finetuning.  }{\includegraphics[width=0.23\textwidth]{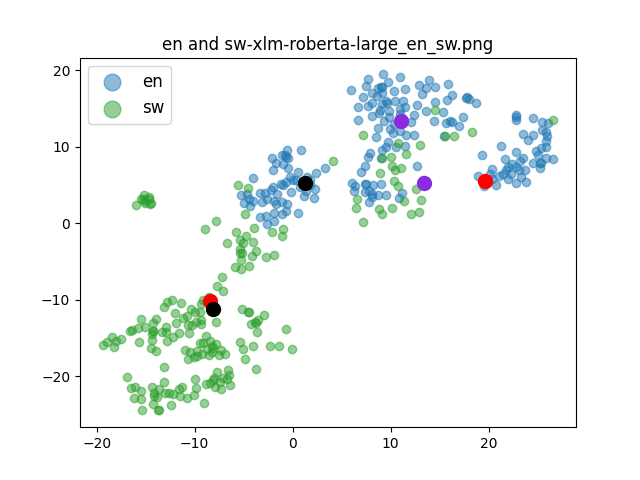}}%
\hfill
\subcaptionbox{XLM-R fintuned on English NLI data.}{\includegraphics[width=0.23\textwidth]{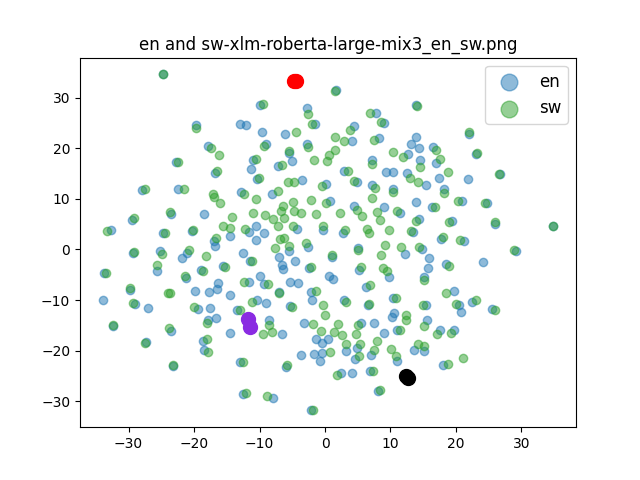}}%
\hfill
\caption{We visualize the sentence embeddings on XNLI corpus, where blue dots and green dots denote the sentences from English and Swahili respectively. Here, red dots, black dots, and  purple dots denote the parallel sentences from different languages. In (a), the sentence embeddings from different languages are clearly separated into two clusters. In (b), after English NLI training, the embedding space becomes indistinguishable for different languages, and the parallel sentences are aligned to each other.}
\label{fig:vis}
\end{figure}

As shown in Figure~\ref{fig:vis} (a), without finetuning on downstream tasks, the embedding space of pre-trained multilingual language models such as m-BERT~\cite{devlin-etal-2019-bert} or XLM-R~\cite{conneau-etal-2020-unsupervised} separate the embeddings of each language into different clusters.
To align cross-lingual sentence embeddings, previous work~\cite{Artetxe_2019,chidambaram-etal-2019-learning,https://doi.org/10.48550/arxiv.2007.01852} finetunes multilingual language models with billions of parallel data.
However, it is non-trivial to obtain numerous parallel data for all languages.
One potential direction to alleviate the need for parallel data is to enhance cross-lingual transfer of sentence embeddings.

Pre-trained multilingual language models~\cite{pires-etal-2019-multilingual,phang-etal-2020-english} have shown impressive performance on cross-lingual zero-shot transfer~\cite{pires-etal-2019-multilingual} that a model finetuned on a source language can generalize to target languages. 
This implies the representations finetuned on downstream tasks are universal across various languages.
In this work, we explore various cross-lingual transfer settings on sentence retrieval tasks, especially in the setting of using English data only.


We propose multilingual-SimCSE (mSimCSE) which extends SimCSE~\cite{gao2021simcse}, a famous sentence embedding method on English, to multilingual for cross-lingual transfer.  
SimCSE is a contrastive learning~\cite{1467314,10.1109/CVPR.2006.100,chen2020simple} method that pulls closer semantically similar sentences (i.e. positive sentence pairs) in embeddings space.
As done in SimCSE, we obtain positive training pairs by either natural language inference (NLI)~\cite{ conneau-etal-2017-supervised,reimers-gurevych-2019-sentence} supervision or unsupervised data augmentation using dropout.
We also investigate model performance when a small amount of parallel data or cross-lingual NLI data are available.

In our experiments, as shown in Figure~\ref{fig:vis} (b), we are surprised to find that contrastive learning on pure English data seems to be able to align cross-language representations.
Sentences that are semantically similar across languages are clearly closer together.
Compared with previous unsupervised or weakly supervised methods, our unsupervised method significantly improves the performance on cross-lingual STS and sentence retrieval tasks.
In retrieving low-resource languages and STS tasks, our method is even on par with fully supervised methods trained on billions of parallel data.
Our results show that using contrastive learning to learn sentence relationships is more efficient than using massively parallel data for learning universal sentence embeddings.
To the best of our knowledge, we are the first to demonstrate that using only English data can effectively learn universal sentence embeddings.

\begin{figure*}[th!]
\centering
\includegraphics[width=0.85\textwidth]{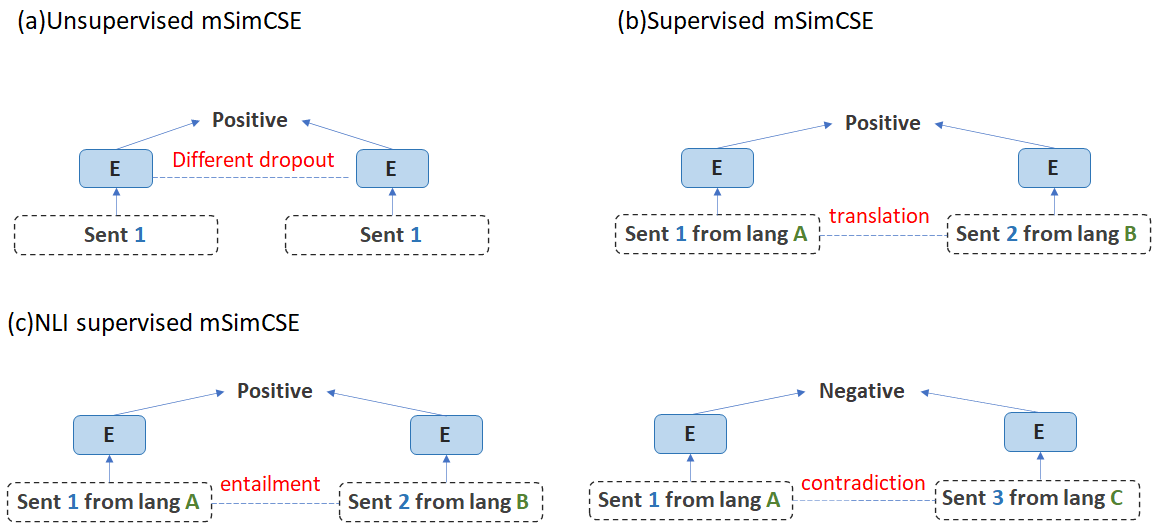}
\caption{
Overview of our method. In (a) unsupervised mSimCSE, sentence 1 is from English Wikipedia. It uses different dropout masks at each encoder inference as data augmentation. In (b) supervised SimCSE, we use parallel sentences as a positive training pair. In (c) NLI supervised mSimCSE, for the model only leverages English data, languages A, B, and C are all English. For mSimCSE that uses cross-lingual NLI supervision, languages A, B, and C are randomly sampled from a language pool. We use entailment and contradiction relationships between sentences to construct positive and hard negative training pairs.}
\label{fig:method}
\end{figure*}

\section{Related Work}
\subsection{Cross-lingual Zero-shot Transfer Learning}
Due to the similarities between different languages, such as words, grammar, and semantics, multilingual models~\cite{devlin-etal-2019-bert,conneau2019cross,conneau-etal-2020-unsupervised,wei2021on,chi-etal-2022-xlm} have been shown to generalize to unseen languages in a wide range of tasks.
In machine translation, multilingual models exhibit the ability to perform zero-shot transfer in that they can translate on unseen language pairs~\cite{zoph-etal-2016-transfer,johnson2017google}.
With the help of self-supervised learning, model can better acquire language-invariant representation, thus improving zero-shot machine translation~\cite{siddhant-etal-2020-leveraging,liu2020multilingual}.

To learn language-invariant representations cross-lingual tasks, previous work apply adversarial networks~\cite{keung-etal-2019-adversarial,chen-etal-2019-multi-source} or align representations via parallel corpus~\cite{Cao2020Multilingual}.
~\citet{pires-etal-2019-multilingual} revealed that mBERT is good at zero-shot cross-lingual transfer that fintuning on a specific monolingual task can generalize to other languages.
More recently, X-Mixup~\cite{yang2022enhancing}  performs manifold mixup of source and target languages to learn general representations.
Our method is the most relevant to the work which enhances cross-lingual transfer by using English downstream tasks to finetune multilingual language models~\cite{phang-etal-2020-english} in that both methods leverage English NLI supervision.
Our method differs from theirs in that we apply contrastive learning for representation learning and we only focus on sentence embeddings.

\subsection{English Sentence Embeddings}
Sentence embeddings aim to map sentences into a shared embedding space in which sentences with similar meanings can be close to each other.
Previous work learns sentence embeddings by predicting the surrounding sentences of an input sentence, in either generative~\cite{NIPS2015_f442d33f} or discriminative~\cite{logeswaran2018an} manners.
Recently, with the success of contrastive learning on learning visual representations~\cite{https://doi.org/10.48550/arxiv.2002.05709}, more and more work explores contrastive learning on sentence embeddings.
The training signal of contrastive learning can be obtained by data augmentation~\cite{fang2020cert,yan-etal-2021-consert,NEURIPS2021_c2c2a045,carlsson2021semantic} or self-guided architecture~\cite{kim-etal-2021-self,carlsson2021semantic}.
Among these, SimCSE~\cite{gao2021simcse} is the most famous one, which adopts either NLI supervision to define relevant sentences or unsupervised dropout as data augmentation, and improves state-of-the-art results.
We choose to extend SimCSE to multilingual due to its impressive performance and simplicity.

\subsection{Cross-lingual Sentence Embeddings}
Universal cross-lingual sentence embeddings align semantically similar cross-lingual sentences into a shared embeddings space.
To learn cross-lingual embeddings, previous work uses large amounts of parallel data to train neural networks to bring the embeddings of parallel sentences closer.
LASER~\cite{artetxe2019massively} trains Bi-LSTM with parallel sentences of 93 languages to encourage consistency between the cross-lingual sentence embeddings.
MUSE~\cite{yang-etal-2020-multilingual} learns universal sentence embedding for 16 languages via translation based bridge tasks including multifeature question-answer prediction, translation ranking, and NLI.
LaBSE~\cite{https://doi.org/10.48550/arxiv.2007.01852} leverages dual-encoder framework to finetune mBERT on 6 billion parallel sentence pairs over 109 languages.
\citet{reimers-gurevych-2020-making} extend the monolingual sentence embedding to multilingual by teacher-student training, in which a target language student model mimics source language embeddings from the teacher model.
Sentence piece encoder (SP)~\cite{wieting2021paraphrastic} simply learns sentence piece embeddings using parallel data and outperforms BERT-base models.

It is also possible to learn cross-lingual sentence embeddings in weakly supervised or even unsupervised manners.
CRISS~\cite{NEURIPS2020_1763ea5a} uses the representations of mBART encoder as initial sentence embeddings, which are used to mine parallel texts from monolingual corpora.
They iteratively extract bitexts and use the bitexts to update the model in a self-training loop.
Notice that our method can be easily combined with CRISS by using mSimCSE as the initial model. 
Our method is closely related to DuEAM~\cite{goswami2021cross}, a dual-encoder model that leverages word mover distance and cross-lingual parallel NLI sentences to construct positive training pairs.
The main differences are that we use contrastive learning, which is more effective than dual encoder, and that our method is more simple that only requires English data.

\vspace{-1mm}
\section{Method}
\vspace{-2mm}
\subsection{SimCSE}
SimCSE~\cite{gao2021simcse} apply batch contrastive learning~\cite{https://doi.org/10.48550/arxiv.2002.05709} to learn sentence embeddings on English data.
Batch contrastive learning puts positive training pairs and negative training pairs into a same batch, increasing the difficulty of a contrastive task.

Specifically, a positive training pair means the sentences in the pair are semantically similar and their embeddings should be pulled closer, while a negative training pair means the sentences are semantically different.
Given a batch of positive training pairs $B=\{(x_i,x_i^+)\}_{i=1}^N$, SimCSE calculates the batch constrastive loss for i-th pair as:
\begin{equation}
    \mathcal{L}_i =  -\log{\frac{e^{sim(h_i,h_i^+)}}{\sum_{j=1}^{N} e^{sim(h_i,h_j^+)}}}.
\end{equation}
Here, $N$ denotes the batch size,
$(x_i,x_i^+)$ denotes two semantically similar sentences, and $h_i = E(x_i)$ is the sentence embedding from encoder $E$.
The key to using this loss is how to define the semantically similar pairs, which we elaborate on in the following section.

\subsection{Multilingual SimCSE}
In this section, we elaborate on how we extend SimCSE to multilingual and illustrate our proposed mSimCSE in Figure~\ref{fig:method}.
We explore four different multilingual training strategies, including the unsupervised strategy, the English NLI supervised strategy, the parallel NLI supervised strategy, and the fully supervised strategy.
The difference between different strategies is how to define a positive training pair.
Here, both unsupervised and English NLI supervised strategies can be recognized as an ``unsupervised'' setting for multilingual training because both of them only use English data and do not use any parallel data.

\paragraph{Unsupervised mSimCSE}
In unsupervised SimCSE, $x_i$ and $x_i^+$ are the same sentence. As $x_i$ and $x_i^+$ are encoded by the same encoder but with different dropout, the dropout can be viewed as a light-weight data augmentation method.
We use the wikipedia data from the original SimCSE repository to train our model.

\paragraph{English NLI supervised mSimCSE}
In the English NLI supervised strategy, we use English natural language inference (NLI)~\cite{ conneau-etal-2017-supervised,reimers-gurevych-2019-sentence} datasets to construct positive and hard negative training pairs.
Specifically, if two sentences are labeled as ``entailment'' relationship, they are viewed as a positive pair $(x_i$, $x_i^+)$.
For each $x_i$, we also include a hard negative example $x_i^-$ in the same training batch, where $x_i$ and $x_i^-$ are labeled as ``contradiction'' relationship.

\paragraph{Cross-lingual NLI supervised mSimCSE}
The English NLI supervision can be easily extended to the multilingual strategy by constructing a positive training pair from different languages.
We use XNLI~\cite{conneau2018xnli}, which translates English NLI to multiple languages.
Similar to the English NLI strategy, the cross-lingual sentence pairs with ``entailment'' and ``contradiction'' relationship are viewed as positive and negative pairs respectively, but as shown in Figure~\ref{fig:method}, the language of $x_i$, $x_i^+$ and $x_i^-$ are randomly sampled. They can come from either different languages or the same language.

\paragraph{Supervised mSimCSE}
In supervised mSimCSE, we simply define a positive training pair as the parallel sentences from different languages.
This strategy is the same as previous supervised methods, but we use relatively few parallel sentences.
Note that different strategies can be easily combined by mixing training pairs from different strategies.

\begin{table*}[ht]
\begin{center}
\small
\begin{tabular}{cccc}
    \hline
    Models & BUCC & Tatoeba-14 & Tatoeba-36\\
    \hline
    \multicolumn{4}{c}{\textbf{Unsupervised}}\\
    \hline
    XLM-R & 66.0 & 57.6 & 53.4\\
    INFOXLM & - & 77.8 & 67.3\\
    DuEAM & 77.2 & - & -\\
    XLM-E & - & 72.3 & 62.3\\
    HiCTL & 68.4 & - & 59.7\\
    $mSimCSE_{en}$ & 87.5 & 82.0 & 78.0\\
    \hline
    \multicolumn{4}{c}{\textbf{English NLI supervised}}\\
    \hline
    ~\citep{phang-etal-2020-english} & 71.9 & - & 81.2\\
    $mSimCSE_{en}$ & \textbf{93.6} & \textbf{89.9} & \textbf{87.7}\\
    \hline
    \multicolumn{4}{c}{\textbf{Cross-lingual NLI supervised}}\\
    \hline
    $mSimCSE_{en,fr}$ & 94.2 & 90.8 & 88.8 \\
    $mSimCSE_{en,fr,sw}$ & 94.3 & 93.3 & 90.3 \\
    $mSimCSE_{all}$ & \textbf{95.2} & 93.2 & 91.4 \\
    DuEAM & 81.7 & - & - \\
    \hline
    \multicolumn{4}{c}{\textbf{Fully Supervised}}\\
    \hline
    LASER & 92.9 & 95.3 & 84.4\\
    LaBSE & 93.5 & 95.3 & 95.0 \\
    $mSimCSE_{sw}$ & 86.8 & 87.7 & 86.3\\
    $mSimCSE_{fr}$ & 87.1 & 87.9 & 85.9\\
    $mSimCSE_{sw,fr}$ & 88.8 & 90.2 & 88.3\\
    $mSimCSE_{sw,fr}$+NLI & 93.6 & 91.9 & 90.0\\
    \hline
\end{tabular}
\end{center}
\caption{Results of sentence retrieval task on Xtreme benchmark. We report F1-scores for BUCC and accuracy for Tatoeba.}
\label{tab:sent_retrieval}
\end{table*}

\begin{table*}[ht]
\begin{center}
\small
\begin{tabular}{ccccccccccc}
    \hline
    Models & hi & fr & de & af & te & tl & ga &  ka & am & sw\\
    \hline
    \multicolumn{11}{c}{\textbf{Unsupervised}}\\
    \hline
    CRISS & 92.2 & 92.7 & 98.0 & - &  - & - & - & - & - & -\\
    DuEAM & 83.5 & - & 93.4 & 79.9 & 78.6 & 56.8 & 35.0 & 70.7 & 46.4 & -\\
    $mSimCSE_{en}$ & $86.9$ & $87.2$ & $94.1$ & $76.0$ & $78.8$ & $49.7$ & $39.2$ & $75.2$ & $48.8$ & $29.4$\\
    \hline
    \multicolumn{11}{c}{\textbf{English NLI supervised}}\\
    \hline
    $mSimCSE_{en}$ & \textbf{94.4} & \textbf{93.9} & \textbf{98.6} & \textbf{85.6} & \textbf{92.9} & \textbf{70.0} & \textbf{54.8} & \textbf{89.2} & \textbf{79.5} & \textbf{42.1}\\
    \hline
    
    \hline
    \multicolumn{11}{c}{\textbf{Cross-lingual NLI supervised}}\\
    \hline
    DuEAM & 92.9 & - & 96.0 & 84.8 & 90.6 & 60.6 & 42.0 & 76.4 & 56.0 & -\\
    $mSimCSE_{en,fr}$ & $95.1$ & $94.4$ & $98.8$ & $88.9$ & $94.2$ & $73.4$ & $59.4$ & $91.3$ & $79.5$ & $44.5$\\
    $mSimCSE_{en,fr,sw}$ & $95.7$ & $94.2$ & $98.4$ & $87.9$ & $94.4$ & $75.6$ & $62.1$ & $90.5$ & $82.7$ & \textbf{75.5}\\
    $mSimCSE_{all}$ & \textbf{96.2} & $94.8$ & $98.8$ & \textbf{90.6} & \textbf{96.2} & \textbf{80.9} & \textbf{65.1} & \textbf{92.4} & \textbf{82.4} & $67.8$\\

    \hline
    \multicolumn{11}{c}{\textbf{Fully supervised}}\\
    \hline
    LASER & 94.7 & 95.7 & 99.0 & 89.4 & 79.7 & - & 5.2 & 35.9 & 42.0 & 42.4\\
    $mSimCSE_{sw}$ & $94.3$ & $91.6$ & $97.6$ & $85.2$ & $88.5$ & $76.3$ & $60.8$ & $85.5$ & $65.2$ & $47.6$ \\
    $mSimCSE_{fr}$ & $94.1$ & $92.6$ & $97.3$ & $84.6$ & $89.3$ & $70.8$ & $54.6$ & $86.3$ & $63.4$ & $43.6$ \\
    $mSimCSE_{sw,fr}$ & $95.1$ & $93.8$ & $97.8$ & $86.1$ & $91.2$ & $75.8$ & $59.6$ & $88.9$ & $74.4$ & $51.5$\\
    $mSimCSE_{sw,fr}$+NLI & $95.8$ & $94.7$ & $98.6$ & $89.8$ & $95.7$ & $77.8$ & $63.9$ & $91.7$ & $81.0$ & $57.1$\\
    \hline

\end{tabular}
\end{center}
\caption{Accuracy of Tatoeba multilingual retrieval task.}
\label{tab:tatoeba}
\end{table*}

\section{Experiments}
\subsection{Experimental Setup}
\paragraph{Training Details}
We adapt SimCSE codebase~\footnote{\url{https://github.com/princeton-nlp/SimCSE}} to a multi-lingual setting.
We keep all other hyperparameters same as the original SimCSE, and fix learning rate to be $1e-5$, training epoch to be $1$, and batch size to be $128$ for all experiments.
We use our method to finetune XLM-Roberta-large (XLM-R)~\cite{conneau-etal-2020-unsupervised}.
We examine the performance of different hyperparameters in Appendix~\ref{sec:app_hyper}

\paragraph{Training Data for Different mSimCSE Strategies}
In unsupervised mSimCSE and English NLI supervised mSimCSE, we use the pre-processed English Wikipedia and English NLI training tuples downloaded from the SimCSE codebase respectively.
In all the tables in this paper, the subscripts of $mSimCSE$ denote the languages that we use to train our model.
In cross-lingual NLI supervision, $mSimCSE_{en,fr}$ denotes we use English and translated French NLI data to train our model and $mSimCSE_{all}$ means that we use all the languages in XNLI~\cite{conneau2018xnli} dataset. 

In supervised finetuning, $mSimCSE_{sw}$ denotes that we use the translation pairs of English and Swahili.
For each language, we randomly select 100k parallel sentences from ParaCrawl project~\footnote{\url{ http://paracrawl.eu}} via the OPUS corpus collection~\footnote{\url{https://opus.nlpl.eu/}}.
In supervised finetuning, ``$mSimCSE_{sw}$+NLI'' denotes that we mix English NLI sentence pairs with English-Swahili translation pairs.
Note that because parallel sentences don't have hard negative sentences, to mix them with NLI data, we also remove hard negative sentences of NLI in ``$mSimCSE_{sw}$+NLI''.

\subsection{Baselines}
First, we compare our method to unsupervised pre-trained language models including XLM-R and M-BERT without finetuning to show that unsupervised contrastive learning can learn more generalized cross-lingual sentence embeddings.
In some tasks, we also compare our method with more competitive language models in Xtreme~\cite{hu2020xtreme} benchmark, such as XLM-E~\cite{chi-etal-2022-xlm}, HICTL~\cite{wei2021on} and INFOXLM~\cite{chi-etal-2021-infoxlm}.
CRISS~\cite{NEURIPS2020_1763ea5a} is an unsupervised sentence retrieval method, which mines parallel sentences from multiple monolingual corpora using self-training.

Our main baselines are other methods which also leverage NLI supervision.
~\citet{phang-etal-2020-english} finetune multilingual language models on various English tasks, including English NLI task.
DuEAM~\cite{goswami2021cross} also leverages multilingual NLI supervision to learn universal sentence embeddings.

We also compare our method with fully supervised methods that leverages parallel sentences, including LASER~\cite{artetxe2019massively}, LaBSE~\cite{https://doi.org/10.48550/arxiv.2007.01852}, and SP~\cite{wieting2021paraphrastic}.
Note that among all the methods, $mSimCSE_{en}$ is the only method that only uses English data.

\subsection{Sentence Retrieval}
Following the previous work of cross-lingual sentence embedding learning~\cite{goswami2021cross}, we evaluate our model on multi-lingual sentence retrieval, including Tatoeba~\cite{artetxe2019massively} and BUCC~\cite{zweigenbaum2018overview}. 
Tatoeba requires models to match parallel sentences from source and target language sentence pools.
BUCC is a bitext mining task, in which a model needs to rank all the possible sentence pairs, and predicts sentence pairs whose scores are above an optimized threshold. 

In Table~\ref{tab:sent_retrieval}, we follow the setting in Xtreme benchmark to evaluate model performance on sentence retrieval task.
In unsupervised setting, mSimCSE trained on English Wikipedia improves the performance by a large margin.
This implies that contrastive training can effectively pull closer cross-lingual semantically similar sentences.

With English NLI supervision, it significantly improves the performance against unsupervised methods and DuEAM.
It even beats fully-suprvised methods that leverages parallel sentences on BUCC task.
This implies that with an objective that learns more difficult semantic relationship between sentences, model can learn better universal cross-lingual sentence embeddings.

By comparing cross-lingual NLI supervised mSimCSE with Enlgish NLI supervised mSimCSE, we observe that the model performance can be further improved using translated NLI pairs from other languages.
In general, including more languages can improve the performance.
Comparing with DuEAM which also leverages parallel NLI supervision, contrastive learning can learn universal sentence embedding more effectively.
In the BUCC dataset, the performance of $mSimCSE_{all}$ is better than fully supervised methods, such as LaBSE, which is trained on 6 billion parallel data.
Note that $mSimCSE_{all}$ uses far less parallel data than LaBSE, which demonstrates the effectiveness of our method.

In the fully supervised setting, comparing $mSimCSE_{sw,fr}$ with and without NLI supervision, we find that if the translation pairs are rare, adding English NLI supervision can significantly improve the performance.
Also, compared with the mSimCSE that only uses English NLI supervision, adding a few extra parallel data can slightly improve the performance.

Following DuEAM, in Table~\ref{tab:tatoeba}, we select some high-resource languages including Hindi (hin), French (fra), German
(deu), Afrikaans (afr) and Swahili (swh), and low-resource languages including Telugu (tel), Tagalog (tgl),
Irish (gle), Georgian (kat), and Amharic (amh) from Tatoeba dataset to further analyze the model performance.
In high-resource languages, fully-supervised achieves better performance because large amounts of parallel sentence pairs are available in these languages.
On the other hand, in low-resource languages, due to the lack of training pairs, supervised method can not generalize well on these languages while mSimCSE can generalize better.

Finally, we evaluate whether including cross-lingual NLI supervision in a target language can improve the performance.
In Table~\ref{tab:tatoeba}, compared with using only English NLI supervision, $mSimCSE_{en,fr,sw}$ in the cross-lingual NLI setting which includes Swahili in training significantly improves the performance in Swahili.
Its performance gain is greater than fully supervised ``$mSimCSE_{sw,fr}$+NLI'', which leverages parallel sentences.

\begin{table}[ht]
\begin{center}
\scriptsize
\begin{tabular}{cccccc}
    \hline
    Models & ar-ar & ar-en & es-es & es-en & tr-en\\
    \hline
    \multicolumn{6}{c}{\textbf{Unsupervised}} \\
    \hline
    XLM-R & 53.5 & 26.2 & 68.1 & 10.7 & 10.5\\
    mBERT & 55.2 & 28.3 & 68.0 & 23.6 & 17.3\\
    $mSimCSE_{en}$ & 72.3 & 48.4 & 83.7 & 57.6 & 53.4\\
    \hline
    \multicolumn{6}{c}{\textbf{English NLI supervised}}\\
    \hline
    $mSimCSE_{en}$ & \textbf{81.6} & 71.5 & \textbf{87.5} & \textbf{79.6} & 71.1\\
    \hline
    \multicolumn{6}{c}{\textbf{Cross-lingual NLI supervised}}\\
    \hline
    DuEAM & 69.7 & 54.3 & 78.6 & 56.5 & 58.4\\
    $mSimCSE_{all}$ & 79.4 & 72.1 & 85.3 & 77.8 & 74.2\\
    \hline
    \multicolumn{6}{c}{\textbf{Fully Supervised}}\\
    \hline
    LASER & 79.7 & - & 57.9 & - & 72.0\\
    LaBSE & 80.8 & - & 65.5 & - & 72.0\\
    SP & 76.7 & 78.4 & 85.6 & 77.9 & 79.5\\
    $mSimCSE_{sw,fr}$+NLI & 77.7 & 72.4 & 86.3 & 79.7 & 72.5\\
    \hline
    
\end{tabular}
\end{center}
\caption{ Spearman rank correlation ($\rho $) results for SemEval 2017 STS shared task. The results of supervised baselines and DuEAM are taken from ~\citep{goswami2021cross}.}
\label{tab:multists}
\end{table}

\subsection{Cross-lingual STS}
Cross-lingual STS~\cite{cer2017semeval} evaluates whether a model predicted semantic similarity between two sentences are correlated to human judgement.
The two sentences can come from either the same language or different languages.
Given a sentence pair, we compute the cosine similarity of sentence embeddings as a model prediction.

The results of multi-lingual STS benchmark are shown in Table~\ref{tab:multists}.
For unsupervised XLM-R and mBERT without finetuning, we try several pooling methods and find that averaging over the first and the last layers yields the best results on STS.
The poor results of pre-trained language models mean that the sentence embeddings of pre-trained language models do not capture the semantics in the cosine similarity space well.
With unsupervised mSimCSE pre-training, it enhances the semantics of monolingual STS tasks, i.e. ``ar-ar'' and ``es-es''.
For the tasks that requires cross-lingual alignment, including ``ar-en'', ``en-es'', and ``tr-en'', the gap between unsupervised baselines and English NLI supervised baselines is still large.

Comparing with the methods that utilize either parallel NLI supervision or supervised parallel data, the English NLI training achieves the best results on ar-ar, es-es, and es-en pairs.
This implies that mSimCSE can effectively transfer semantic knowledge learned from English NLI supervision to other languages.
We find that using parallel data can only improve the results of bilingual pairs, and reduces the performance of monolingual pairs.

\subsection{Unsupervised Classification}
We conduct unsupervised classification to evaluate whether the model can cluster semantically similar documents together on the languages other than English.
We use Tnews dataset in CLUE benchmark~\footnote{\url{https://github.com/CLUEbenchmark/CLUE}} to evaluate the performance of our model.
Tnews is a Chinese news classification dataset, which contains 15 news categories.
We first conduct k-means clustering on sentence embedding, in which cluster number $k$ is set to be the same as the number of news categories.
Then, we evaluate the mean accuracy of each cluster that measures what percentages of documents in each cluster are from the same human-labeled category.

Compared with unsupervised pre-trained language models, mSimCSE significantly improves the purity scores.
This is expected because without fine-tuning, the embeddings from pre-trained language models cannot capture relative distances between instances.
Similar to the observation in the previous section, English NLI supervision can greatly enhance the performance, closing the gap between the fully supervised fine-tuned BERT.

\begin{table}[ht]
\begin{center}
\small
\begin{tabular}{cc}
    \hline
    Models & Purity\\
    \hline
    \multicolumn{2}{c}{\textbf{Unsupervised}}\\
    \hline
    Random & 6.7 \\
    mBERT & 15.2 \\
    XLM-R & 13.7 \\
    $mSimCSE_{en}$ & 30.3\\
    \hline
    \multicolumn{2}{c}{\textbf{English NLI supervision}}\\
    \hline
    $mSimCSE_{en}$ & \textbf{40.3}\\
    \hline
    \multicolumn{2}{c}{\textbf{Cross-lingual NLI supervision}}\\
    \hline
    $mSimCSE_{all}$ & 41.6 \\
    \hline
    \multicolumn{2}{c}{\textbf{Supervised Classification Model}}\\
    \hline
    BERT & 56.6\\
    \hline
\end{tabular}
\end{center}
\caption{Accuracy of unsupervised clustering on Tnews classification dataset. In supervised fintuning, the model is finetuned on classification training set.}
\label{tab:clustering}
\end{table}

\subsection{Zero-shot Cross-lingual Transfer of Sentence Classification}
To evaluate the cross-lingual zero-shot transfer of pre-trained sentence embedding, we evaluate our model on PAXS-X~\cite{Yang2019paws-x} sentence classification task. 
PAXS-X requires a model to determine whether two sentences are paraphrases.
In Table~\ref{tab:pawsx}, compared with XLM-R and XLM-E~\cite{chi-etal-2022-xlm}, mSimCSE without using NLI data improves the performance, which demonstrates that mSimCSE is an effective approach for zero-shot cross-lingual transfer.
In this task, using English NLI supervision does not improve performance.

\begin{table}[ht]
\begin{center}
\small
\begin{tabular}{cc}
    \hline
    Models & Accuracy\\
    \hline
    \multicolumn{2}{c}{\textbf{Unsupervised}} \\
    \hline
    mBERT & 81.9\\
    XLM-R & 86.4\\
    XLM-E & 87.1\\
    $mSimCSE_{en}$ & \textbf{88.1}\\
    \hline
    \multicolumn{2}{c}{\textbf{English NLI supervised}}\\
    \hline
    ~\citep{phang-etal-2020-english} & 87.9\\
    $mSimCSE_{en}$  & 88.2\\
    \hline
\end{tabular}
\end{center}
\caption{Accuracy on PAWS-X dataset.}
\label{tab:pawsx}
\end{table}

\section{Analysis}
\subsection{The Effect of Parallel Sentences Number}
\begin{table}[ht]
\begin{center}
\small
\begin{tabular}{cccc}
\hline
Parallel data & BUCC & Tatoeba14 & Tatoeba36\\
\hline
0 & 91.4 & 90.4 & 88.0\\
10k & 92.6 & 90.6 & 88.5\\
100k & 93.5 & 90.8 & 88.6\\
1M & 94.4 & 90.6 & 88.5\\
5M & 94.5 & 90.7 & 88.2\\
\hline
\end{tabular}
\end{center}
\caption{The effect of parallel English-French sentences number.}
\label{tab:ana_parallel}
\end{table}

As parallel data is easy to obtain for most languages~\cite{artetxe-etal-2020-call}, we investigate the effect of the number of parallel sentences.
We mix parallel English-French sentences with the English NLI data and gradually increase the number of parallel sentences.
Here, 0 parallel sentence means we only use English NLI data without using hard negative examples, so the results are different from the results in Table~\ref{tab:sent_retrieval}.

The BUCC dataset has only four high-resource languages, of which French is one of them.
With more parallel data, the consistent improvement on BUCC dataset implies that using more English-French translation pairs can improve the performance on the English-French mining task, thus improving the results of BUCC.
On the other hand, the Tatoeba dataset includes much more languages, which evaluates a model's generalization.
We observe that using more parallel data does not influence the performance on Tatoeba, which implies that using translation data on a single language pair does not generalize well to other languages.
The results suggest that using large amounts of parallel data may not be the most efficient way to learn universal sentence embeddings while learning sentence relationships is a more promising direction.

\subsection{Can Contrastive Learning Removes Language Identity?}
\begin{table}[ht]
\begin{center}
\small
\begin{tabular}{ccc}
    \hline
    Models & en,de,fr,hi & en,tr,ar,bg\\
    \hline
    XLM-R & 99.2 & 99.8\\
    $mSimCSE_{en}$ & 91.1 & 95.3\\
    \hline
\end{tabular}
\end{center}
\caption{Accuracy of language classifier.}
\label{tab:lg_classify}
\end{table}
Masked language modeling requires a model to capture the language identity to predict correct tokens for a specific language. 
On the other hand, English contrastive loss only learns the relationship between sentences, which does not seem to require language identity.
We speculate that the contrastive loss can thus remove the language identity in sentence embeddings and enhance the general shared cross-lingual semantics. 

To verify this, in Table~\ref{tab:lg_classify}, we train two language classifiers on en,de,fr,hi and en,tr,ar,bg respectively.
The language classifier needs to predict the correct language of the input sentence embeddings.
We use the sentences from XNLI as our training and testing data.
With contrastive learning, the accuracy of language classifier decreases, which implies the embeddings are more language-invariant, which more or less verifies our assumption.
However, the accuracy is still very high because the language classifier can still predict the language of a text by language-specific features such as grammar and characters.

\section{Discussion}
Our experimental results demonstrate that in both unsupervised and English NLI supervised settings, using English data alone can surprisingly align cross-lingual sentence embeddings.
By comparing unsupervised results with NLI supervised results, we observe that learning more meaningful sentence relationships can further enhance the alignment.
In our analysis, we find that infinitely increasing parallel training data is not the most efficient manner to learn universal sentence embeddings;
instead, our results suggest that designing a more challenging contrastive task or more effective sentence embedding learning method on English data may be a more efficient direction.
Also, contrastive learning may be a promising direction for improving the zero-shot transfer of pre-trained multilingual language models.

We attribute the alignment to language-invariant contrastive training.
Because multilingual language models have shown good performance on zero-shot transfer, we speculate that multilingual language models encode texts into two disentangled embeddings, a language-specific embedding and a general language agnostic embedding. 
Because English contrastive task doesn’t require mlms to capture language identity, it only pulls closer language-agnostic sentence embeddings while weakening language-specific embedding. 
This property can be verified in Figure~\ref{fig:vis} and Table~\ref{tab:lg_classify}.
However, it still requires more investigation to fully understand why contrastive learning on English data can achieve cross-lingual transfer.

\section{Conclusion}
In this work, we demonstrate that using only English data can effectively learn universal sentence embeddings.
We propose four different strategies to extend SimCSE to multilingual, including unsupervised, English NLI supervised, cross-lingual NLI supervised, and supervised strategies.
We surprisingly find that the English NLI supervised strategy can achieve performance on par with previous supervised methods that leverage large amounts of parallel data.
Our work provides a new perspective on learning universal sentence embeddings and cross-lingual transfer that using contrastive learning to learn sentence semantic relationships on monolingual corpora may be a promising direction.

\section{Limitations}
In the previous sections, we attribute why models trained on English data can learn cross-lingual sentence embeddings to language-invariant contrastive task. 
We speculate that multilingual language models have already implicitly learned such universal representations, but they also learn some language-specific representations.
Contrastive learning enhances the language-invariant representations, diminishing the language-specific representations without distorting the semantics embedded in the representations.
However, this speculation still requires more evidence to support it.
Also, it is important to understand to which the zero-shot transfer happens, such as which languages are easier to transfer, and what is the properties of these languages.
For the properties of these languages, by observing the experimental results in Table~\ref{tab:tatoeba}, we have two speculations, one is their similarity between English, and another one is the number of the monolingual pre-training data for these languages, but again these speculations also require more analysis to verify.
By understanding the reason for this phenomenon, it is possible to achieve better zero-shot transfer and learn more ``universal'' sentence embeddings.


\bibliography{emnlp2022}
\bibliographystyle{acl_natbib}

\appendix
\clearpage

\begin{table} [h!] \small
\centering
\begin{tabular}{ccc|ccc}
\hline
epoch & bs & lr & BUCC & Tatoeba-14 & Tatoeba-36\\
\hline
1 & 128 & 1e-5 & 93.6 & 89.9 & 87.7\\
2 & 128 & 1e-5 & 94.4 & 90.0 & 87.8\\
3 & 128 & 1e-5 & 93.5 & 90.0 & 87.4\\
1 & 256 & 1e-5 & 93.3 & 90.0 & 87.9 \\
1 & 128 & 2e-5 & 93.7 & 90.0 & 87.4\\
\hline
\end{tabular}
\caption{Performance of different Hyperparameters. Epoch denotes training epoch number, bs denotes the batch size, and lr denotes learning rate.}
\label{table:hyper}
\end{table}

\section{Hyperparameters} \label{sec:app_hyper}
In Table~\ref{table:hyper}, we show how different hyperparameters influence the model performance. We choose $mSimCSE_{en}$ trained on English NLI data as the model under examination.
We find that the performance of different hyperparameters is very close, which implies that our method is stable and not sensitive to hyperparameters.
Increasing the number of training epochs to 2 can  improve the performance on BUCC.


\end{document}